\documentclass[11pt]{article}

\usepackage[margin=1in]{geometry}
\usepackage{newtxtext}
\usepackage{amsmath}
\usepackage{amssymb}
\usepackage{newtxmath}
\usepackage{graphicx}
\usepackage{booktabs}
\usepackage{natbib}
\usepackage{hyperref}
\hypersetup{hidelinks}
\usepackage{url}
\title{SegBench-GC: Testing Segmentation Invariance in Multi-Step Offline Goal-Conditioned Reinforcement Learning}
\author{Musa Shams\\Independent Researcher\\\href{https://orcid.org/0009-0005-1015-5342}{ORCID: 0009-0005-1015-5342}\quad\href{https://github.com/MusaShams/SegBench-GC}{github.com/MusaShams/SegBench-GC}}
\date{August 2026}

\begin{document}
\maketitle

\begin{abstract}
Offline goal-conditioned reinforcement learning (GCRL) often uses trajectory structure for future-goal sampling and multi-step targets, yet logged trajectories may be partitioned for administrative reasons that do not correspond to termination. We introduce SegBench-GC, a controlled stress test of \emph{segmentation invariance} that holds transitions, source trajectories, goal sampling, optimization settings, and evaluation fixed while varying only artificial backup boundaries and whether those boundaries retain continuation value. Continuation-valid targets (CVT) provide the segmentation-consistent control: reward accumulation stops at an artificial cut, but the target bootstraps from its stored successor. In a matched-count PointMaze study with 35,000 artificial cuts, three segmentation realizations, and three optimization seeds, final 50-episode-per-task success is 50.5\% uncut, 39.1\% with CVT, and 19.1\% when the same cuts are treated as absorbing; across segmentation realizations, naive mean success ranges from 4.8\% to 31.9\%. An independent published $n$-step baseline ($n=25$) from the Decoupled Q-Chunking codebase shows the same failure on Puzzle-4x5: 47.2\% uncut, 58.5\% CVT, and 0.27\% naive across three optimization seeds. A target-level diagnostic verifies $y_{\mathrm{naive}}-y_{\mathrm{CVT}}=-\gamma^k V(s_{t+k},g)$ to numerical precision, and learned-critic diagnostics show a large optimistic shift under naive handling while CVT remains approximately aligned with the uncut critic. CVT applies standard continuation bootstrapping rather than a new Bellman rule; the contribution is the controlled benchmark, failure isolation, and cross-learner evidence that administrative segmentation can materially change multi-step offline GCRL.
\end{abstract}

\section{Introduction}

Offline goal-conditioned reinforcement learning (GCRL) promises reusable control policies from fixed datasets without task-specific online interaction. Modern methods exploit trajectory structure to sample future goals and construct multi-step targets, making episode boundaries part of the learning problem rather than inert metadata \cite{ogbench2024}.

Those boundaries need not represent true task termination. Data collectors split streams at time limits, storage limits, worker restarts, or preprocessing boundaries, and multi-step learners commonly train from fixed windows or chunks whose endpoints truncate a return even when the underlying process continues. Treating such administrative cuts as absorbing states removes continuation value from multi-step targets. Correct bootstrapping at nonterminal truncations is established theory \cite{pardo2018timelimits}, but the sensitivity of offline GCRL to arbitrary segmentation has not been isolated while holding transitions, source-trajectory semantics, future-goal sampling, optimization settings, and evaluation fixed.

We study \emph{segmentation invariance}: a trajectory-aware offline GCRL result should be stable, up to optimization noise, when identical logged transitions are partitioned at different nonterminal locations. SegBench-GC separates source-trajectory boundaries used for goal sampling from artificial \emph{backup} boundaries used only to truncate a multi-step target. This separation prevents resegmentation from silently changing the goal distribution or the underlying logged process.

Our contributions are:
\begin{itemize}
  \item a controlled segmentation-invariance protocol for offline GCRL in which transition tuples, source trajectories, sampled-goal semantics, optimization settings, and evaluation are fixed while artificial backup boundaries are varied;
  \item a continuation-valid control that preserves standard nonterminal bootstrapping at artificial cuts while leaving genuine goal completion noncontinuing;
  \item a matched-count PointMaze study over three optimization seeds and three independent segmentation realizations showing a large performance penalty and strong segmentation sensitivity when only the artificial cuts are treated as absorbing;
  \item target- and critic-level mechanism diagnostics, together with an independent replication using the published $n$-step ($n=25$) baseline from the Decoupled Q-Chunking codebase \cite{li2026dqc}.
\end{itemize}

Empirically, artificial-terminal handling reduces primary mean success by 20.0 percentage points relative to CVT and increases segmentation dispersion from 3.4 to 13.6 points. In the independent Puzzle-4x5 validation, the same intervention reduces mean success from 58.5\% with CVT to 0.27\%. These results support a deliberately scoped claim: multi-step offline GCRL can be sensitive to administrative segmentation when backup boundaries are treated as terminals. CVT is not proposed to outperform correctly uncut data; it is the continuation-consistent control for this intervention.

\section{Related Work}

\paragraph{Truncation and timeout semantics.}
Time limits used to facilitate training should not be conflated with terminal states of the underlying Markov decision process. Pardo et al.\ formalize this distinction and show that partial episodes require value bootstrapping at truncation \cite{pardo2018timelimits}. Contemporary environment APIs distinguish termination from truncation for the same reason, and D4RL similarly exposed timeout metadata in offline datasets \cite{fu2020d4rl}. CVT applies this known principle to trajectory-aware multi-step offline GCRL; the novelty of SegBench-GC is the controlled invariance benchmark, not the Bellman correction itself.

\paragraph{Offline goal-conditioned RL.}
OGBench evaluates offline GCRL under stitching, long-horizon reasoning, mixed-quality data, and high-dimensional observations \cite{ogbench2024}. Its datasets distinguish trajectory ends from Bellman masks: trajectory boundaries constrain future-goal sampling, while masks indicate task success. SegBench-GC preserves that distinction and adds a separate set of artificial backup boundaries, allowing target sensitivity to be measured without changing the goal distribution.

\paragraph{Temporal abstraction and multi-step value propagation.}
Implicit Q-learning and conservative Q-learning learn from fixed datasets without online exploration \cite{kostrikov2022iql,kumar2020cql}. More recent offline-GCRL work directly targets long effective horizons. Horizon Reduction studies several ways to shorten the learning horizon, including n-step GCSAC+BC \cite{park2025horizon}; OTA incorporates option-aware temporal abstraction into value learning \cite{ahn2025ota}; multistep quasimetric learning combines multistep propagation with quasimetric structure \cite{zheng2026mqe}; Decoupled Q-Chunking separates critic and policy chunk lengths while retaining multi-step value propagation \cite{li2026dqc}; and Transitive RL replaces linear-depth TD propagation with a divide-and-conquer value update \cite{park2026trl}. In their reported evaluations, these methods do not use benign resegmentation as a counterfactual reliability intervention while holding transitions and future-goal semantics fixed.

\paragraph{Segmentation and information loss.}
Recent theory shows that restricting offline RL to fixed-length trajectory segments can cause identifiability failure, objective misspecification, and support aliasing \cite{nidadala2026horizon}. SegBench-GC studies a complementary setting: ordered transitions remain available, source trajectories remain intact for goal sampling, and only a separate backup segmentation is changed. Hierarchical action-chunking work likewise uses $k$-step value backups for long-horizon offline GCRL \cite{jawaid2026hiqc}, but does not isolate whether an administrative backup cut retains continuation value. We therefore treat SegBench-GC as a diagnostic protocol for learners that consume trajectory prefixes while limiting empirical claims to the two tested learners.

\paragraph{Reliable empirical evaluation.}
Deep RL comparisons are highly variable and should report paired seeds, uncertainty, and distributions rather than isolated point estimates \cite{agarwal2021statistical}. We pair optimization seeds and segmentation configurations where possible, aggregate segmentation variants within seed before aggregating across seeds, and report paired contrasts alongside descriptive variability.

\section{Method}

Let $\mathcal{D}$ contain ordered transitions $(s_t,a_t,s_{t+1})$ and source-trajectory boundaries $\tau_t$. Source boundaries constrain future-goal sampling and describe how the logged stream was collected. In OGBench, task completion is represented separately by a Bellman mask. SegBench-GC introduces a second object, an artificial backup-boundary set $b$, that truncates a multi-step return without changing any transition tuple or source-trajectory boundary.

\subsection{Controlled resegmentation}

Artificial cuts are inserted only at eligible nonterminal locations with a valid stored successor. Source trajectories $\tau$ remain fixed and continue to define future-goal sampling, so two runs that differ only in $b$ see the same underlying transitions and goal-sampling semantics. The Original control ignores artificial cuts entirely. In the segmented conditions, reward accumulation stops when a cut is encountered; CVT and naive handling differ only in whether continuation value is retained at that same cut. In the primary matched-count experiment we sample an exact number of artificial cuts with a seeded procedure; in the independent validation we match the same cut density. This design isolates backup-boundary semantics from changes in data content, future-goal support, or evaluation.

\subsection{Continuation-valid targets}

For a requested backup horizon $h$, let $k\leq h$ be the number of valid steps before goal completion or the first artificial backup cut. At an artificial cut whose stored successor belongs to the same continuing process, CVT uses
\begin{equation}
  y_{\mathrm{CVT}}^{(h)}
  =
  \sum_{i=0}^{k-1}\gamma^i r_{t+i}
  + \gamma^k V^{(h)}(s_{t+k},g).
\end{equation}
The naive artificial-terminal control uses the \emph{same cut} but sets only that artificial-cut continuation to zero:
\begin{equation}
  y_{\mathrm{naive}}^{(h)}
  =
  \sum_{i=0}^{k-1}\gamma^i r_{t+i}.
\end{equation}
Source-trajectory boundaries are continuation-valid in both conditions, and genuine goal completion disables bootstrap in both conditions. Thus the CVT-versus-naive intervention is isolated to artificial cuts. At any affected artificial cut,
\begin{equation}
  y_{\mathrm{naive}}^{(h)}-y_{\mathrm{CVT}}^{(h)}
  =-\gamma^k V^{(h)}(s_{t+k},g).
\end{equation}
For the negative goal reward used by the primary learner, continuing values are almost always non-positive, so dropping the continuation term shifts the naive target upward. Administrative cuts can therefore become optimistic pseudo-terminals even though the underlying process continues.

CVT is not a new Bellman rule. It is the continuation-consistent control obtained by applying standard nonterminal truncation semantics at an artificial boundary whose successor is known to be valid \cite{pardo2018timelimits}. Under a continuing fixed-point interpretation, Bellman expansion makes the cut target agree with the corresponding uncut target at consistency. An explicitly finite-horizon value would instead require a residual-horizon quantity such as $V^{(h-k)}$.

\subsection{Segmentation invariance}

For algorithm $A$, fixed dataset transitions $\mathcal{D}$, and artificial segmentation $b$, let $J_A(b)$ denote rollout performance of the learned policy. Let $b_0$ be the uncut condition and let $\mathcal{B}$ be a set of artificial segmentations. We summarize sensitivity with
\begin{align}
  \mathrm{SegGap}(A)
  &=J_A(b_0)-\frac{1}{|\mathcal{B}|}\sum_{b\in\mathcal{B}}J_A(b),\\
  \mathrm{WorstDrop}(A)
  &=J_A(b_0)-\min_{b\in\mathcal{B}}J_A(b),\\
  \mathrm{SegDisp}(A)
  &=\operatorname{Std}_{b\in\mathcal{B}}\left[J_A(b)\right].
\end{align}
When multiple segmentations are evaluated for one optimization seed, we first aggregate over segmentations within that seed and then aggregate across optimization seeds. This avoids treating segmentation realizations as independent training replicates.

\subsection{Test subjects and goal protocol}

Our primary test subject is a goal-conditioned IQL-style learner with twin horizon-indexed critics and value heads for $\{1,2,4,8\}$ \cite{kostrikov2022iql}. The actor uses a fixed soft mixture of horizon-conditioned primitive-action heads. Because these heads are trained from multi-step targets, their learning signal depends on backup-boundary semantics. One-step GCIQL is an exact negative control: when multi-step horizon targets are disabled, artificial backup-boundary metadata is not consumed and paired replay samples are identical under resegmentation.

For cross-implementation validation, we use the published NS baseline from the Decoupled Q-Chunking codebase \cite{li2026dqc}, configured with policy chunk size one, backup horizon $25$, and no chunk critic. This is a conventional multi-step $n$-step learner rather than our horizon-indexed implementation, providing a separate test of whether the failure is implementation-specific.

Primary PointMaze goals follow the official OGBench protocol \cite{ogbench2024}. Value goals mix current, future-trajectory, and random states with probabilities $0.2/0.5/0.3$; actor goals mix future and random states with probabilities $0.5/0.5$. Goal completion receives reward zero and disables bootstrap regardless of segmentation.

\section{Experiments}

\subsection{Primary matched-count study}

The primary study uses OGBench PointMaze Medium Stitch \cite{ogbench2024}. The learner has three 512-unit GELU layers with layer normalization, expectile $0.9$, batch size 1024, learning rate $3\times10^{-4}$, discount $0.99$, and DDPG+BC coefficient $\alpha=0.003$. We train for 100k updates with optimization seeds $\{0,1,2\}$.

For each optimization seed, we evaluate three independently seeded artificial segmentations, $\{101,102,103\}$. Each segmentation contains exactly 35,000 artificial cuts sampled from eligible nonterminal locations, approximately 3.5\% of the eligible PointMaze stream. CVT and naive runs use the same cut set for each pair of optimization and segmentation seeds. The uncut Original condition has no artificial cuts. Final results use 50 episodes per evaluation task, or 250 rollouts per checkpoint.

OGBench collection-trajectory ends are not task completion events. In all three conditions we hold source-boundary continuation semantics fixed and require a finite stored successor at every continuing source boundary. The only CVT-versus-naive difference is the bootstrap mask at an \emph{artificial} cut: CVT retains continuation, whereas naive handling zeros it. This isolates the intervention to the artificial backup boundaries.

\subsection{Independent published $n$-step validation}

To test whether the effect is specific to our horizon-indexed learner, we use the published NS baseline from the Decoupled Q-Chunking codebase \cite{li2026dqc} on Puzzle-4x5 Play. We use the paper's NS configuration with two critics, policy chunk size one, backup horizon $25$, no chunk critic, mean Q aggregation, expectile distillation, quantile implicit backup, $\kappa_b=0.7$, and batch size 4096. Training uses 250k updates. The training budget was selected from an Original-only learnability check before CVT or naive outcomes were examined.

The independent validation uses optimization seeds $\{100001,200002,300003\}$ and one fixed artificial segmentation realization (seed 101) at the same 3.5\% cut density as the primary study. Pairing is verified directly: with the NumPy sampling state reset, CVT and naive minibatches are identical in every key except the artificial-cut continuation mask. Final reevaluation again uses 50 episodes per task. Because this validation uses one fixed segmentation realization, it provides evidence for cross-learner boundary-semantic sensitivity rather than a second segmentation-dispersion study.

\subsection{Diagnostics and negative control}

We use two mechanism diagnostics on the primary learner. First, a direct target calculation samples 32,768 replay starts and constructs CVT and naive targets from the same transitions and the same trained CVT value function, isolating the algebraic effect of the artificial-cut continuation mask. We check the identity
$y_{\mathrm{naive}}-y_{\mathrm{CVT}}=-\gamma^kV(s_{t+k},g)$
only on target slots affected by artificial cuts. Second, for all nine optimization/segmentation pairs we compare learned value and Q predictions at distances zero through eight transitions from the next artificial cut, using matched states and deterministically sampled future goals.

One-step GCIQL serves as an exact negative control because it does not consume the artificial backup-boundary metadata used by the multi-step target constructor. Paired replay checks therefore require its sampled observations, goals, rewards, and masks to remain identical under artificial resegmentation.

\subsection{Metrics and aggregation}

The primary metric is mean success over the five fixed evaluation tasks. For the primary study, segmentation realizations are averaged within each optimization seed before means and sample standard deviations are computed across optimization seeds. We additionally report segmentation-seed means across optimization seeds to expose sensitivity to the arbitrary cut realization. For the independent validation, sample standard deviations are across the three optimization seeds under the one fixed segmentation realization. With three training seeds, all uncertainty summaries are descriptive rather than well-powered inferential confidence intervals \cite{agarwal2021statistical}.

\section{Results}

\begin{table}[t]
  \centering
  \caption{Final five-task success (\%, mean $\pm$ sample SD over optimization seeds). In the primary study, the three segmentation realizations are averaged within each optimization seed before aggregation. The independent NS validation uses one fixed segmentation realization. The final column reports the paired CVT-minus-naive difference.}
  \label{tab:main-results}
  \begin{tabular}{lrrrr}
\toprule
Setting & Original & CVT & Naive & CVT$-$Naive \\
\midrule
Primary PointMaze, 100k & $50.5\pm15.4$ & $39.1\pm5.1$ & $19.1\pm0.2$ & $+20.0\pm5.4$ \\
Published NS baseline, 250k & $47.2\pm14.3$ & $58.5\pm19.5$ & $0.27\pm0.23$ & $+58.3\pm19.8$ \\
\bottomrule
\end{tabular}

\end{table}

\subsection{Artificial terminalization substantially degrades the primary learner}

In the matched-count PointMaze study, the uncut Original condition reaches $50.5\pm15.4$\% success. CVT reaches $39.1\pm5.1$\%, while treating only the same 35,000 artificial cuts as absorbing lowers success to $19.1\pm0.2$\%. The paired CVT-minus-naive difference is $20.0\pm5.4$ percentage points. This degradation occurs despite identical source-boundary semantics, transitions, goals, architecture, and optimization settings.

CVT is not perfectly invariant. Relative to the uncut control, its mean segmentation gap is 11.4 percentage points, its worst observed segmentation drop is 15.1 points, and the descriptive sample SD across the three segmentation means is 3.4 points. Under naive artificial-terminal handling, the corresponding values are a 31.5-point mean gap, a 45.7-point worst drop, and 13.6-point segmentation dispersion. CVT retains 77.4\% of the uncut mean at this training budget. We therefore treat CVT as the segmentation-consistent control rather than as evidence that all residual segmentation effects disappear in finite-sample learning.

\subsection{The arbitrary segmentation realization materially changes naive performance}

The three segmentation realizations produce CVT means of 39.6\%, 42.3\%, and 35.5\% after averaging optimization seeds. Under naive artificial-terminal handling, the corresponding means are 20.5\%, 31.9\%, and 4.8\%. At the individual optimization/segmentation-run level, naive success ranges from 0\% to 37.6\%.

This dispersion is central to the benchmark: no transition or source trajectory changes across these segmentation draws. The learner changes only because different nonterminal locations are declared to be artificial backup boundaries. A method that is insensitive to administrative segmentation should not exhibit this scale of performance variation from that metadata choice alone.

\subsection{The target-level mechanism is exact and produces learned critic optimism}

The direct target diagnostic evaluates 131,072 horizon-specific target slots from 32,768 sampled replay starts. Artificial cuts affect 12,585 of those slots. On the affected slots, the numerical residual of
\begin{equation}
  y_{\mathrm{naive}}-y_{\mathrm{CVT}}
  =-\gamma^kV(s_{t+k},g)
\end{equation}
is at most $3.8\times10^{-6}$. The naive target is shifted upward on 99.8\% of affected slots, with a mean shift of 39.26. Because both targets use the same sampled transitions and the same trained CVT value function, this diagnostic checks the isolated target construction rather than comparing separately trained critics.

The learned-critic diagnostic shows the same mechanism after training. Averaged over distances zero through eight from artificial cuts and then over all nine optimization/segmentation pairs, naive-minus-CVT value predictions are $+15.54\pm2.13$ and naive-minus-CVT Q predictions are $+15.84\pm1.62$. By contrast, CVT-minus-Original is $-0.07\pm4.14$ for value and $+0.17\pm3.63$ for Q. Artificial-terminal targets therefore induce a broad optimistic value shift, whereas the CVT critic remains approximately centered on the uncut critic in aggregate.

\subsection{A published $n$-step baseline independently reproduces the failure}

The published NS ($n=25$) baseline from the Decoupled Q-Chunking codebase \cite{li2026dqc} reproduces the same qualitative failure outside our horizon-indexed implementation. Final Puzzle-4x5 success is $47.2\pm14.3$\% uncut, $58.5\pm19.5$\% with CVT, and $0.27\pm0.23$\% with naive artificial-terminal handling. The paired CVT-minus-naive difference is $58.3\pm19.8$ percentage points; the three per-seed gaps are 52.0, 80.4, and 42.4 points.

CVT also exceeds the uncut control by 11.3 points on average in these three runs, but we do not interpret that as a general improvement claim. The relevant controlled contrast is that the same fixed artificial segmentation retains substantial task success under continuation-valid targets and nearly eliminates goal-reaching behavior when those cuts are made absorbing.

\subsection{Negative control}

One-step GCIQL is invariant by construction because it does not consume the multi-step backup-boundary metadata. An exact paired replay check confirms that artificial resegmentation leaves its sampled observations, goals, rewards, and masks unchanged. This negative control supports localization of the observed failure to multi-step target construction.

\section{Limitations}

CVT assumes that the successor at an artificial cut is observed and belongs to the same continuing process. It must not bootstrap across a genuine reset, missing-data discontinuity, environment termination, or goal completion. The benchmark therefore applies only when boundary semantics can be identified well enough to distinguish continuation from true termination.

The primary study uses three optimization seeds and three segmentation realizations on PointMaze Medium. The independent validation adds a separate published $n$-step implementation on Puzzle-4x5, but it still uses only three optimization seeds and one fixed segmentation realization. Reported standard deviations are descriptive rather than well-powered inferential confidence intervals. These experiments support a cross-learner reliability failure, not a claim that every multi-step offline RL algorithm will exhibit the same magnitude of sensitivity. Broader validation should include additional algorithm families, manipulation domains with strong uncut performance, and naturally fragmented datasets with documented boundary causes.

CVT does not achieve perfect segmentation invariance in the primary study. At 100k updates it reaches 39.1\% versus 50.5\% for the uncut control, a residual gap of 11.4 percentage points. Approximation error, finite optimization, and the fact that horizon-indexed heads need not satisfy an exact continuing fixed point can all leave residual sensitivity even when boundary semantics are correct. Segmentation invariance should therefore be treated as a measurable reliability objective rather than an assumed binary property.

\section{Conclusion}

SegBench-GC treats trajectory segmentation as an explicit reliability intervention rather than inert dataset metadata. Making artificial nonterminal cuts absorbing substantially reduces performance and makes the primary learner sensitive to the arbitrary segmentation realization. Target- and critic-level diagnostics connect that behavioral degradation to optimistic pseudo-terminal targets, and a separate published $n$-step implementation reproduces the same failure pattern outside our primary learner.

The broader implication is methodological: multi-step offline RL should make boundary semantics explicit and should test sensitivity to benign resegmentation when sequence boundaries enter target construction. CVT is the continuation-consistent control for that test. SegBench-GC contributes a controlled protocol for isolating and measuring this failure mode.

\subsection*{Ethics statement}

This study uses public benchmark data and simulated environments; it involves no human subjects, personal data, or physical deployment. The main methodological risk is misclassifying administrative data boundaries as genuine terminations. We address this risk by stating the required continuation assumptions, releasing condition-level aggregate artifacts and protocol tests, and limiting claims to the tested learner families.

\subsection*{Reproducibility statement}

The matched protocols and aggregation rules appear in Section~4. The appendix provides the target argument, hyperparameters, seed-level outcomes, pinned external revisions, and artifact organization. Supplementary source code provides configurations, runners, deterministic aggregate tables, automated tests, and paper-build and packaging utilities.

\bibliographystyle{iclr2027_conference}
\bibliography{references}

\appendix
\section{Continuation-valid target identity}
\label{app:cvt}

Consider a nonterminal artificial cut after $k<h$ valid transitions. Let $V$ satisfy the continuing Bellman equation along the stored trajectory. The CVT target is
\begin{equation}
  y_{\mathrm{CVT}}
  =
  \sum_{i=0}^{k-1}\gamma^i r_{t+i}
  + \gamma^k V(s_{t+k},g).
\end{equation}
The corresponding artificial-terminal target is
\begin{equation}
  y_{\mathrm{naive}}
  =
  \sum_{i=0}^{k-1}\gamma^i r_{t+i},
\end{equation}
so
\begin{equation}
  y_{\mathrm{naive}}-y_{\mathrm{CVT}}
  =-\gamma^kV(s_{t+k},g).
\end{equation}
Repeatedly expanding a consistent continuing value for the remaining $h-k$ transitions recovers the uncut backup. This identity applies only to a boundary whose successor is part of the same continuing process; genuine goal completion, environment termination, reset discontinuities, and missing successors remain noncontinuing. An explicitly finite-horizon value would require a residual-horizon quantity such as $V^{(h-k)}$.

\section{Primary experimental details}
\label{app:details}

The primary learner uses horizons $\{1,2,4,8\}$, twin critics, horizon-indexed value and actor outputs, three hidden layers of width 512, GELU activations, layer normalization, discount $0.99$, expectile $0.9$, batch size 1024, learning rate $3\times10^{-4}$, target update coefficient $0.005$, and DDPG+BC coefficient $0.003$. The actor executes a fixed horizon mixture with weights $(0.05,0.05,0.05,0.85)$.

PointMaze goals follow the official OGBench mixture: value goals use current, future-trajectory, and random states with probabilities $(0.2,0.5,0.3)$; actor goals mix future-trajectory and random states equally. The source adapter uses non-compacted OGBench transitions so every continuing collection boundary retains its stored successor. The OGBench revision used in the reported experiments is \texttt{1d4140997f60c52c6fb0702ec100dc988b18c548}.

The primary intervention samples exactly 35,000 artificial cuts independently for segmentation seeds 101, 102, and 103. Optimization seeds are 0, 1, and 2. Source boundaries remain continuation-valid in Original, CVT, and naive conditions; at artificial cuts, CVT bootstraps while naive handling does not. Final reevaluation uses 50 episodes for each of five fixed tasks.

\section{Independent NS validation details}
\label{app:independent}

The independent baseline is the published NS configuration from the Decoupled Q-Chunking repository \cite{li2026dqc}, pinned at revision \texttt{df898256a77f3594b54a7268bd5f89915981da35}. We use Puzzle-4x5 Play with two critics, policy chunk size one, backup horizon 25, no chunk critic, expectile distillation, quantile implicit backup, mean Q aggregation, $\kappa_b=0.7$, $\kappa_d=0.5$, and batch size 4096. This preserves the paper's $n$-step learner while avoiding a chunk-critic confound in which post-cut actions would enter the critic input.

Artificial cuts are sampled at 3.5\% of eligible successor-state locations with segmentation seed 101 and are fixed across optimization seeds 100001, 200002, and 300003. CVT and naive minibatch pairing is checked by resetting the NumPy sampling state before each draw; every batch field must match except the continuation mask on cut-affected backups. A 4096-sample verification batch contained 1846 cut-affected backups and passed this equality check. Training uses 250k updates, and final reevaluation uses 50 episodes per task.

\section{Seed-level results}
\label{app:seeds}

\begin{table}[h]
\centering
\small
\begin{tabular}{llrrr}
\toprule
Study & Optimization seed & Original & CVT & Naive \\
\midrule
PointMaze primary & 0 & 32.8 & 33.2 & 19.3 \\
PointMaze primary & 1 & 58.4 & 42.4 & 18.9 \\
PointMaze primary & 2 & 60.4 & 41.7 & 18.9 \\
\midrule
Puzzle NS & 100001 & 44.0 & 52.4 & 0.4 \\
Puzzle NS & 200002 & 62.8 & 80.4 & 0.0 \\
Puzzle NS & 300003 & 34.8 & 42.8 & 0.4 \\
\bottomrule
\end{tabular}
\caption{Seed-level final five-task success (\%). Primary PointMaze CVT and naive entries first average the three segmentation realizations; the Puzzle NS validation uses one fixed segmentation realization.}
\label{tab:seed-results}
\end{table}

Primary segmentation-seed means are Naive: 20.5\%, 31.9\%, and 4.8\%; CVT: 39.6\%, 42.3\%, and 35.5\% for seeds 101, 102, and 103, respectively.

\section{Artifact organization}
\label{app:artifacts}

Every training run writes a JSONL file containing the resolved configuration, optimization seed, boundary counts and continuation assumptions, total update step, held-out evaluation, task-level rollout success, and checkpoint metadata. Final reevaluations, target-identity diagnostics, and boundary-local critic contrasts are stored separately from training logs. The aggregation protocol averages segmentation realizations within an optimization seed before aggregating optimization seeds. Raw checkpoints are excluded from source control because of size; released manifests and generated tables provide provenance for the reported summaries.

\end{document}